\documentclass[10pt,conference]{IEEEtran}
\IEEEoverridecommandlockouts
\usepackage{cite}
\usepackage{amsmath,amssymb,amsfonts}
\usepackage{algorithmic}
\usepackage{graphicx}
\usepackage{textcomp}
\usepackage{xcolor}
\usepackage{makecell}

\usepackage{array}
\usepackage{makecell}
\usepackage{stfloats}
\usepackage{url}
\usepackage{verbatim}
\usepackage{balance}
\usepackage{booktabs}
\usepackage{multirow}
\usepackage{soul}
\usepackage{algorithmic}
\usepackage[ruled]{algorithm2e}
\usepackage{multirow}
\usepackage{booktabs}
\usepackage{tabularx,lipsum}
\usepackage{caption}
\usepackage{subcaption}
\usepackage[final,commandnameprefix=ifneeded]{changes}

\def\BibTeX{{\rm B\kern-.05em{\sc i\kern-.025em b}\kern-.08em
    T\kern-.1667em\lower.7ex\hbox{E}\kern-.125emX}}
\begin{document}

\title{\Huge BicKD: Bilateral Contrastive Knowledge Distillation%\\[0.1cm]
\thanks{
J. Zhu, Y. Xu and Y. Gu are with Kyushu University, Fukuoka, Japan.  
% $^{\dagger}$%Joint Graduate School of Mathematics for Innovation, 
% Kyushu University, Fukuoka, Japan 
% $^{*}$Faculty of Information Science and Electrical Engineering, Kyushu University, Japan
%$^{\ddagger}$%Department of Computer Science, 
%
L. Xiong and H. Lee are with 
Emory University, Atlanta, USA. 
Y. Liu is with Beijing University of Technology, Beijing, China. 
J. Liu is with The Hong Kong Polytechnic University, Hong Kong.
% $^{\S}$
% $^{\dagger\dagger}$

J. Zhu and Y. Xu contributed equally to this work. 
J. Zhu was supported in part by the WISE program (MEXT) at Kyushu University in this work.
(email: zhu.jiangnan.584@s.kyushu-u.ac.jp)
}
}
%}

\author{
Jiangnan Zhu, %$^{\dagger}$, 
Yukai Xu, %$^{\dagger}$,
Li Xiong, %$^{\ddagger}$, 
Yixuan Liu, %$^{\S}$,
Junxu Liu, %$^{\dagger\dagger}$,
Hong kyu Lee, %$^{\ddagger}$,
Yujie Gu %$^{\dagger}$
% \IEEEauthorblockN{1\textsuperscript{st} Given Name Surname}
% \and
% \and
}

\maketitle

\begin{abstract}
Knowledge distillation (KD) is a machine learning framework that transfers knowledge from a teacher model to a student model. 
The vanilla KD proposed by Hinton et al. has been the dominant approach in logit-based distillation and demonstrates compelling performance. However, it only performs sample-wise probability alignment between teacher and student's predictions, lacking a mechanism for class-wise comparison. Besides, vanilla KD imposes no structural constraint on the probability space. 
In this work, we propose a simple yet effective method\deleted[]{ology}, bilateral contrastive knowledge distillation (BicKD). This approach introduces a novel bilateral contrastive loss, which intensifies the orthogonality among different class generalization spaces while preserving consistency within the same class. 
The bilateral formulation enables explicit comparison of both sample-wise and class-wise prediction patterns between teacher and student. By emphasizing probabilistic orthogonality, BicKD further regularizes the geometric structure of the predictive distribution.
Extensive experiments show that our BicKD method enhances knowledge transfer, and consistently outperforms state-of-the-art knowledge distillation techniques across various model architectures and benchmarks.
% As a result, BicKD  during distillation process.
% \added[]{However, existing knowledge distillation methods usually show severe class-wise accuracy discrepancies in student model.
% }
% Motivated by this observation, we first conduct experiments to evaluate the range in student models.
% % {\color{red}Statistical performance, such as variance ...}\added[]{class-wise distillation performance using standard deviation over all classes.
% In our experiments, for almost all cases, the
% range across class-wise accuracy exceeds 40\%, 
% ratio of the standard deviation to mean accuracy exceeds 15\%,
% indicating a substantial difference across classes.
% }
% effects of \added[]{this class-wise accuracy discrepancies} on knowledge distillation. 
% Our findings \added[]{further} reveal that \added[]{the large class-wise distillation accuracy gaps negatively affect the overall model accuracy. Some classes show low prediction accuracy, which limits the overall performance, primarily due to differences in learnability among classes.}
% enhances the generalizability of \added[]{classes with low learnability by leveraging the representational strengths of classes are relatively easy to learn, thereby improving the overall model accuracy.}
\end{abstract}


\section{Introduction}
\label{sec:intro}
Nowadays deep neural networks (DNNs) have achieved significant advancements in real-world applications, such as computer vision \cite{imagenet,yolov3}, natural language processing \cite{bert,gpt3}, and speech recognition \cite{LAS,wav2vec2}, etc. 
Large DNN models trained on vast datasets often perform better but incur high storage and computational costs, making it difficult to be deployed on portable devices.

\textit{Knowledge distillation} (KD) \cite{KD} is a machine learning framework used to transfer knowledge from a large high-performing teacher model to a simpler efficient student model. 
%The teacher model is typically well pre-trained using a large dataset. 
The distillation process aims to train the student model to mimic teacher's predictions using a \deleted[]{transfer }dataset. It enables the student model to achieve performance comparable to teacher model while using fewer resources.

\begin{figure}[t]
    \centering
    % \begin{subfigure}[c]{0.48\linewidth}
        % \vspace{0.3cm}
        \centering
        \includegraphics[height=4.2cm,
  width=\linewidth,keepaspectratio]
  % {ICASSP/orthography.pdf}
  {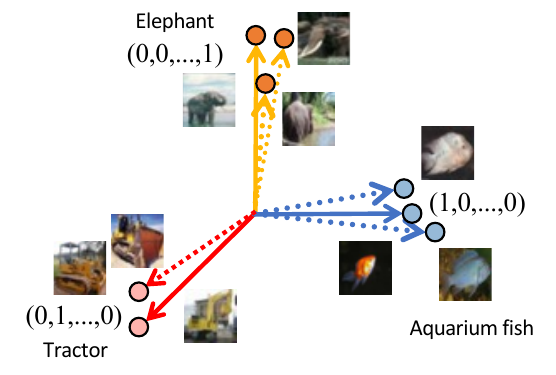}
% \vspace{0.20cm}
        % \caption{Conceptual description.}
  %   \end{subfigure}
  %   \begin{subfigure}[c]{0.48\linewidth}
  %       \centering
  %       \includegraphics[height=6cm,
  % width=\linewidth,keepaspectratio]{ICASSP/orth_bickd.pdf}
  % % {figures/orthogonality_boxed.pdf}
  %       \caption{Empirical 3D visualization.}
  %   \end{subfigure}
    \caption{Class-wise orthogonality in probability space. Ideally, the mean vectors of different classes are orthogonal, and sample-wise predictions are distributed close to the mean vector of each class.
    % (a) shows the predicted probability vectors of three CIFAR-100 classes (elephant, tractor, and aquarium fish), where the vectors of different classes point toward mutually orthogonal directions, corresponding to one-hot target vectors such as $(1,0,\dots,0)$, $(0,1,\dots,0)$, and $(0,0,\dots,1)$. 
    % (b) visualizes the prediction vectors of samples from the first three classes on CIFAR-100 in 3D probability space, where each point corresponds to a sample.
    % Class-mean vectors of different classes are approximately orthogonal, while sample vectors cluster around their corresponding class means. 
    }
    \label{fig:orth}
\end{figure}

However, most existing approaches lack mechanisms for effective comparison across different classes. For instance, the vanilla KD\cite{KD} only aligns teacher and student's outputs of corresponding classes, which make it fails to fully exploit the knowledge cross different classes. In addition, prior methods rarely impose structural constraints on the geometry of the probability space. In the ideal case, the predicted probability distributions of different classes are approximately orthogonal (see Fig.~\ref{fig:orth}). This geometric property implies that different classes occupy well-separated directions in the probability space. Orthogonality-based constraints have been studied in DNN learning to improve training stability and reduce feature redundancy among learned representations\cite{orth}. Motivated by this observation, we introduce an orthogonality-based constraint into knowledge distillation. This constraint aims to regularize the geometric structure of the probability space and promote clearer separation among class predictions. Importantly, the orthogonality enforced here does not simply push the student toward sharper or overly confident outputs. It is designed to mimic the geometric structure learned by the teacher, serving as a teacher-guided structural bias in the distillation process.

To that end, 
in this paper, we introduce a new effective knowledge distillation method, termed \textit{Bilateral Contrastive Knowledge Distillation} (BicKD), 
to promote knowledge distillation through orthogonality-driven regularization while introducing an explicit cross-class comparison mechanism.

%%%%%%%%%%%%%%%%%%%%%%%%%%%%%%%%%%%%%%%%%%%%%%%%%%
\begin{table*}[h]
    \centering
    \tabcolsep=0.25cm
    \caption{Summary on related knowledge distillation methods}
    \begin{tabular}{c c c c c c c c c c c c c}
    \toprule
    Method  & KD & VID & FitNets & RKD & SP & CRD & DIST & MLD & RLD & 
    % LSKD &
    WKD-L & WTTM & \textbf{BicKD} \\
    \midrule
    feature/logits-based 
    & logits & feature & feature & feature& feature & feature & logits & logits & logits & 
    % logits & 
    logits & logits & logits \\
    \midrule
    reference 
    & \cite{KD} & \cite{VID} & \cite{fitnets} & \cite{relationalKD} & \cite{SP} & \cite{CRD} 
    & \cite{dist} & \cite{multi-level-logits}
    & \cite{rld2024} 
    % \cite{lskd2024}  
    & \cite{wkd2024}&\cite{wttm2024} 
    & This work \\
    \bottomrule
    \end{tabular}
\label{tab:baselines}
\end{table*}
%%%%%%%%%%%%%%%%%%%%%%%%%%%%%%%%%%%%

%In particular
Specifically, BicKD introduces bilateral contrast from two aspects:
(1) Sample-wise contrast, which strives to enhance the \emph{orthogonality} between teacher's prediction on one sample and student's prediction on another sample with a different label, \replaced[]{and in the meanwhile preserve alignment between those predictions on identical samples}{and in the meanwhile preserve the relations between those predictions on identical samples}; 
(2) Class-wise contrast, which aims to amplify the \textit{orthogonality} between teacher and student's predictions on distinct classes and to minimize the divergence between their predictions on the same class (see Fig. \ref{fig:rowCol}). 
By emphasizing orthogonality, BicKD 
% enables leveraging the strengths of strong classes to help weak classes, 
%\added[]{promotes more distinct directions for different classes in the probability space,} thereby enhancing 
promotes 
the geometric distinction among different classes in the probability space.
By introducing class-wise contrast, BicKD explicitly models the relationship among cross-class prediction patterns, leading to better separation among class predictions.
This bilateral contrast constitutes the foundation of our proposed BicKD method. 
Extensive experiments show that BicKD outperforms existing state-of-the-art knowledge distillation methods across a variety of model architectures and datasets.%, particularly for weak classes.

In summary, the contributions of this paper are as below.
\begin{itemize}
    % \item First, 
    % \added[]{we identify a discrepancy in distillation accuracy across classes even though the dataset is balanced, and conduct experiments on the CIFAR-100 dataset to analyze this phenomenon.}
    \item We propose a simple yet effective logits-based BicKD framework by introducing a bilateral contrastive loss that contrasts teacher and student's predictions both sample-wise and class-wise. This approach amplifies the orthogonality among distinct classes and retains the consistency within identical classes, 
    providing an explicit cross-class comparison mechanism and guiding the distillation process through geometric structural regularization,
    % enabling  the strong classes to help weak classes, 
    thus improving overall distillation performance. 
    \item 
    We conduct comprehensive experiments that show the proposed BicKD method outperforms state-of-the-art knowledge distillation baselines across a variety of models and benchmarks. In addition, ablation studies verify the effectiveness of individual components in BicKD and show its superiority under non-standard data scenarios \footnote{\replaced[]{Code is available at https://github.com/zian0147/BicKD.}{Code will be available upon acceptance.}}.
    %}{\color{red}add ablation studies???}
\end{itemize}

%{\color{red}
The remainder of this paper is organized as follows. Section~\ref{sec:related-work} reviews related work on KD. Section~\ref{sec:Preliminaries} presents the preliminaries. Section~\ref{sec:proposedmethod} describes the proposed method. Section~\ref{sec:experiments} presents the experimental results and comparisons with existing methods, and Section~\ref{sec:ablationstudy} further provides ablation studies. Finally, Section~\ref{sec:conclusion} concludes the paper.

\section{Related work}
\label{sec:related-work}

%%%%%%%%%%%%%%%%%%%%%%%%%%%%%%%%%%%%
\subsection{Knowledge Distillation}
  Existing KD methods can be roughly categorized into two types: logits-based and feature-based \cite{continual-kd-survey}. 
Logits-based methods typically align student and teacher model's logits, which are easy to implement and can be done in a black-box setting where the internal structures of teacher models are unknown. 
In contrast, feature-based methods aim to minimize the divergence on the intermediate feature representations between teacher and student models in higher-dimensional spaces, which requires the internal structures of teacher model and enables student to learn more nuanced knowledge from teacher in a white-box setting.
However, it increases memory/computation costs and the risk of overfitting. 
Some related KD methods are listed in Table~\ref{tab:baselines}.

%The most commonly used logits-based method, known as 
\subsubsection{Logits-based KD}
Vanilla KD, introduced in the seminal work \cite{KD}, is a commonly-used logits-based method. 
It aligns teacher's and student's logits/predictions by minimizing their Kullback-Leibler (KL) divergence per sample.
Later 
%\cite{dist} pointed out that KL divergence aims to let the student exactly recover the teacher's predictions, which can be ineffective when there is a large discrepancy between the teacher and student models. 
%In response, they
\cite{dist} proposed DIST that aims to preserve relations using a correlation-based metric to improve distillation performance when teacher and student have a large discrepancy. 
\cite{multi-level-logits} proposed a multi-level logits distillation method by preserving the similarities from the teacher to student model on sample, batch, and class levels. 
\cite{boosting-kd-via-intra-class-kd} employed an intra-class logits smoothing technique to reduce overconfidence prediction by the teacher model before distillation. \cite{Parameter-Efficient-kd} trained an additional adapter module connecting the shallow and deep layers of teacher model to soften labels and narrow the gap between teacher and student.
RLD \cite{rld2024} dynamically refined teacher logits to eliminate misleading information of teacher while preserving class correlations.
WKD-L \cite{wkd2024} performed cross-category comparison of probabilities based on discrete Wasserstein distance to leverage interrelations among categories.
WTTM \cite{wttm2024} used an inherent Renyi entropy term and a sample-adaptive weighting coefficient to improve generalization and better match teacher’s probability distribution.

\subsubsection{Feature-based KD}
The first feature-based method FitNets \cite{fitnets} aimed to minimize the discrepancy between the middle layers' parameters of teacher and student models.
Following this, VID 
\cite{VID} introduced to transfer knowledge by maximizing the mutual information between teacher and student's feature representations using the variational information maximization technique.
SP \cite{SP} encourages student model to mimic teacher's feature similarities.
%among samples as exhibited by teacher model.
In addition, \cite{attention,CC,CS} used distinct loss functions for maintaining the relations between teacher and student's feature representations in distillation. 
\cite{ijcai_selfkd, trans_selfkd} proposed a hierarchical self-supervision augmented method to learn more meaningful representations in distillation. \cite{skill-kd} used two meta learning networks to extract and predict teacher's behavior from the hidden feature maps. %\cite{stage} proposed a new loss function based on the distribution of features to optimize learning at each distillation substage. 
\cite{structured} aimed to identify foreground pixels and critical feature channels and capture long-range pixel relationships from teacher models.

%%%%%%%%%%%%%%%%%%%%%%%%%%%
\begin{figure*}[t]\centering
    \includegraphics[scale=0.75]{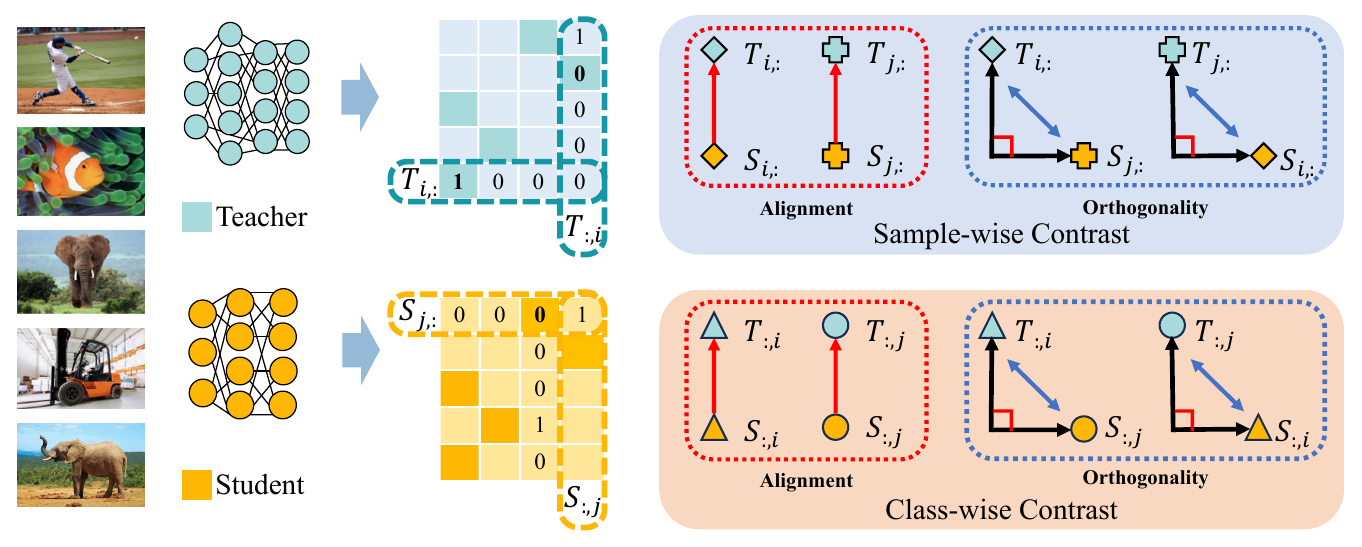}
    \caption{Bilateral contrast used in BicKD. The sample-wise contrast loss is defined to align the same sample/row of teacher and student, and amplify the orthogonality between samples/rows in distinct classes. The class-wise contrast loss is defined to align the same class/column and amplify the orthogonality between different classes/columns of teacher and student models. }
    \label{fig:rowCol}
\end{figure*}
%%%%%%%%%%%%%%%%%%%%%%%%%%%

\subsection{Contrastive Learning}
  Contrastive learning is a type of technique that aligns the feature representations of similar samples while amplifying the disparity between  dissimilar samples, see e.g. \cite{contrastive}. 
In the context of knowledge distillation, CRD 
\cite{CRD} introduced contrastive representation distillation in the way that 
aligning student and teacher's feature representations on identical samples while amplifying the disparity of feature representations on randomly-sampled pairs with different labels. 
However, this method results in additional memory allocation to store feature representations and imposes an increased computational burden due to sampling and contrasting across multiple pairs.
%{\color{red}please clarify the difference between BicKD and CRD, and BicKD can address what kind of challenge, and BicKD can provide something new beyond CRD?}

The proposed BicKD also adopts a contrastive learning paradigm. In contrast to CRD which performs contrastive alignment in the high-dimensional feature space, BicKD performs contrastive alignment on the logits in a much lower-dimensional space.
% that are directly related to the decision outputs. 
This design eliminates the need for additional memory buffer and the extra pair-sampling cost required by CRD, resulting in a lightweight and effective manner during the distillation process. In addition, CRD performs contrastive learning primarily at sample-wise, where the optimization is defined on individual sample pairs. BicKD further introduces class-wise contrast that captures the geometric
structure of sample predictions in the probability space, 
enabling the model to leverage inter-class differences more effectively.
\section{Preliminaries}
\label{sec:Preliminaries}

  Let $B$ and $C$ denote the batch size and the number of classes, respectively. 
In a neural network, denote the output logits as $\mathbf{F}\in \mathbb{R}^{B \times C}$ and the predictions as $\mathbf{P}\in \mathbb{R}^{B \times C}$ which is generated by the \textit{softmax} of logits $\mathbf{F}$ with a temperature factor $\tau$, i.e. 
\begin{equation}
    \mathbf{P}_{i,j} := \frac{e^{\mathbf{F}_{i,j}/\tau}}{\sum_{k=1}^C e^{\mathbf{F}_{i,k}/\tau}}.
\label{softmax}
\end{equation} 
Let $\mathbf{T} \in \mathbb{R}^{B \times C}$ and $\mathbf{S} \in \mathbb{R}^{B\times C}$ denote the predictions of the teacher and student models, respectively. 
%%%%%%%%%%%%%%%%%%%%%%%%%%%%

% \begin{figure*}[h]
%     \begin{subfigure}{0.32\linewidth}
%         \centering
%         \includegraphics[width=\linewidth]{picweakclass_index_evolution_S:resnet20_T:vgg13.pdf}
%         \caption{VGG13 $\rightarrow$ ResNet-20}
%         \label{fig:strong}
%      \end{subfigure}
%     \hfill
%     \begin{subfigure}{0.32\linewidth}
%         \centering
%         \includegraphics[width=\linewidth]{picweakclass_index_evolution_S:resnet8_T:ShuffleV2.pdf}
%         \caption{ShuffleV2 $\rightarrow$ ResNet8}
%         \label{fig:strong}
%      \end{subfigure}
%     \hfill
%     \begin{subfigure}{0.32\linewidth}
%         \centering
%         \includegraphics[width=\linewidth]{picweakclass_index_evolution_S:wrn_16_2_T:resnet110.pdf}
%         \caption{ResNet-110 $\rightarrow$ WRN-16-2}
%         \label{fig:strong}
%      \end{subfigure}
%     \caption{Accuracy curves of the highest class (blue), lowest class (orange), and class mean (grey) of vanilla KD on CIFAR-100. 
%     % The curves depict the evolution of these three types of accuracy during training.
%     }
%     \label{fig:train}
% \end{figure*}

%\subsection{Vanilla Knowledge Distillation}
%\label{subsec:VKD}

  The vanilla KD \cite{KD} aims to align the predictions of student and teacher models via Kullback-Leibler divergence based loss 
$\mathcal{L}_{KL} := \frac{1}{B}\sum^B_{i=1}KL(\mathbf{S}_{i,:},\mathbf{T}_{i,:})$, together with the ground-truth labels based supervised learning 
$\mathcal{L}_{CE} := \frac{1}{B} \sum^B_{i=1}CE(\mathbf{S}_{i,:}, y_i)$,
where $CE$ refers to the cross-entropy function and $y_i$ is the ground truth label of the $i$-th sample. The vanilla KD loss combines $KL$ and $CE$, i.e., 
\begin{align}\label{eq:KD}
    \mathcal{L}^{\text{KD}}_{train} := \lambda \mathcal{L}_{CE} + (1-\lambda) \mathcal{L}_{KL} 
\end{align} 
where $\lambda\in [0,1]$ is a weight for balancing the two parts.

\section{The Proposed Method}
\label{sec:proposedmethod}
  % To \added[]{improve the generalization of weak classes with lower learnability},
  This section presents the proposed bilateral contrastive knowledge distillation (BicKD) framework (see Fig. \ref{fig:rowCol}).

\subsection{Orthogonality in Knowledge Distillation}

  % To strengthen the learning capability of weak classes, our intuition is to let the strong classes help the weak classes 
  %{\color{red}remove weak and strong class???} 

Ideally, the prediction of a sample (row) is  expected to be one-hot, meaning only one class in the prediction is marked as $1$ while the others are $0$ (see Fig.~\ref{fig:orth}). 
Accordingly, the predictions of any two samples in distinct classes are expected to be \emph{orthogonal}, i.e., their inner product is $0$. 
It is worth noting that this orthogonality is expected to hold for all distinct classes. That is, the inner products of the predictions on any two different classes (columns) are supposed to be $0$. 

\begin{figure}
    \centering
    \subfloat[ShuffleNetV2 on CIFAR-100]{
        \includegraphics[width=0.45\linewidth]{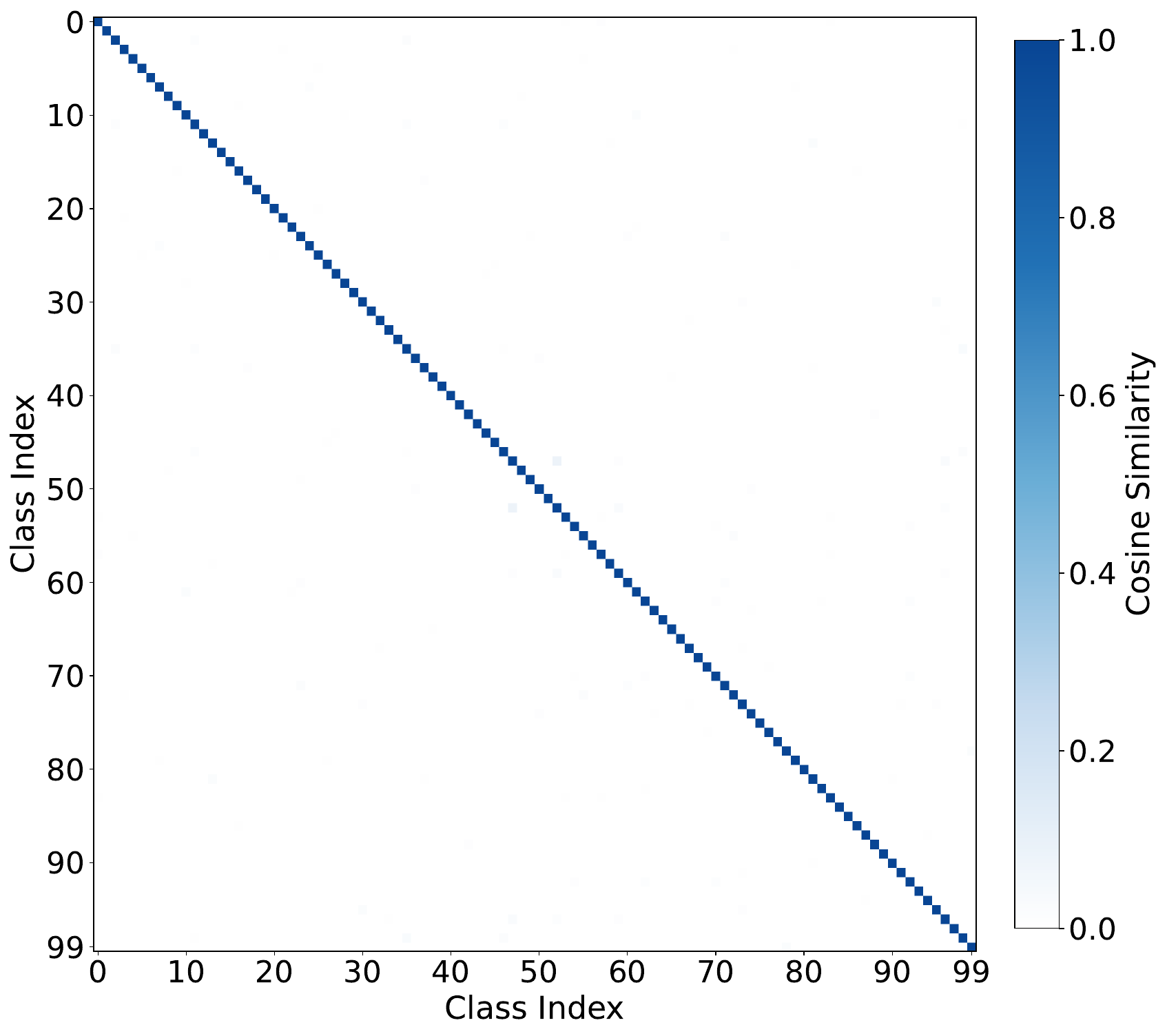}
        \label{fig:sl}
    }
    \hfill
    \subfloat[ResNet-50 on Tiny-ImageNet]{
        \includegraphics[width=0.45\linewidth]{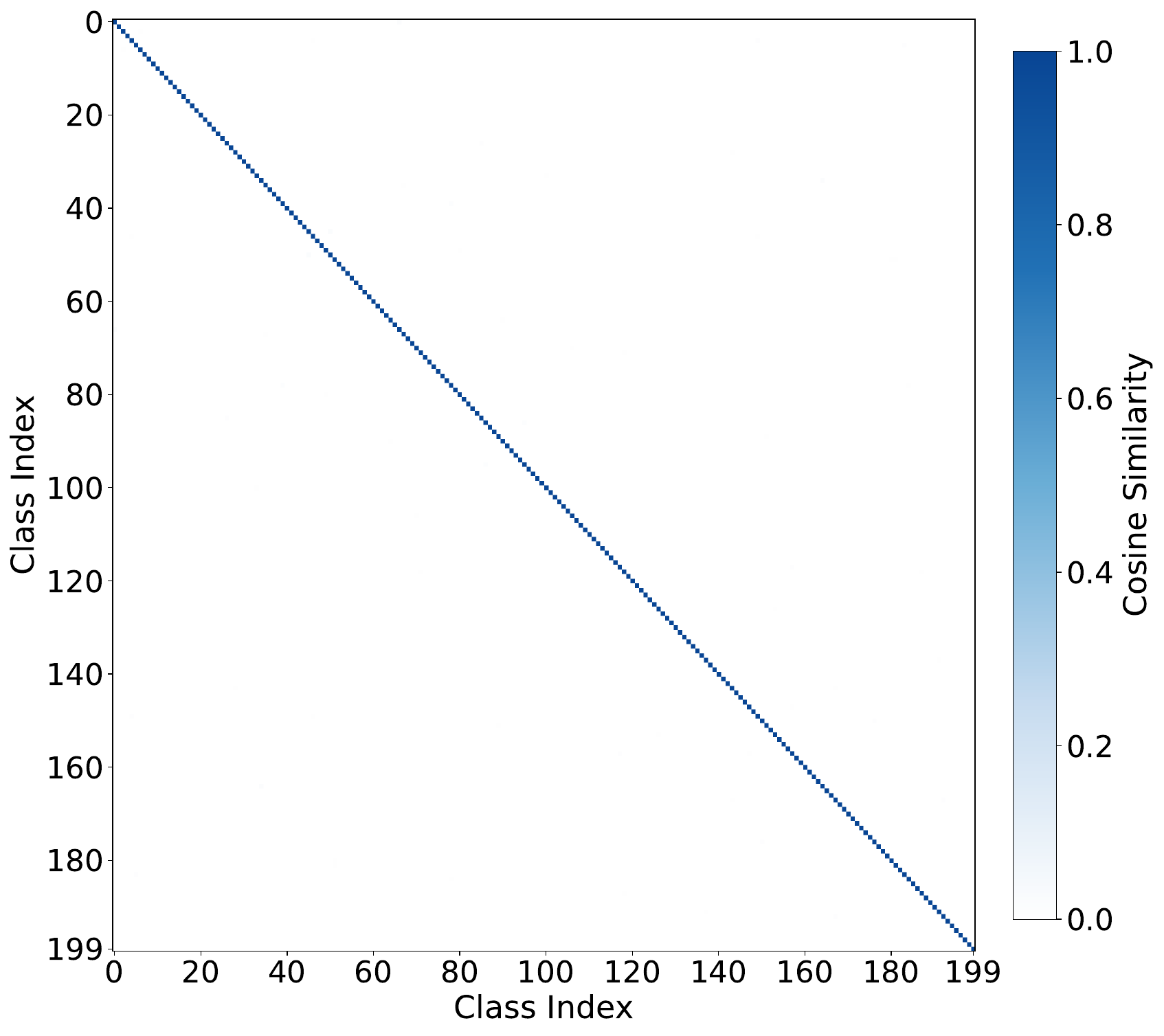}
        \label{fig:cp}
    }
    \caption{Cosine similarity matrices of inter-class predictions of teacher models on the training dataset. For (a), the inter-class (off-diagonal) mean cosine similarity is 0.0004 with std of 0.0018; for (b), the inter-class mean cosine similarity is 0.0001 with std of 0.0004.}
    \label{fig:orthtecher}
\end{figure}

In fact, a DNN's prediction for a specific sample can be viewed as a probability distribution vector. During knowledge distillation, our intuition is to reduce the geometric discrepancy between student's and teacher's predictions in the probability simplex. A well-trained teacher model is expected to achieve high accuracy on the training data and to produce low-entropy, near-vertex outputs, which provide a meaningful geometric reference for the student. 
\added[]{To empirically validate this intuition, we provide the cosine similarity matrices of predictions of teacher models on the training dataset, evaluated on CIFAR-100 \cite{CIFAR100} with ShuffleNetV2 and Tiny-ImageNet \cite{tinyimagenet} with ResNet-50. As shown in Fig.~\ref{fig:orthtecher}, the cosine similarity between predictions of the same class approaches 1, while that between different classes is near 0. Specifically, for ShuffleNetV2 and ResNet-50 respectively, the inter-class mean cosine similarity is 0.0004 and 0.0001, with standard deviations (std) of 0.0018 and 0.0004. These results show that these teacher models exhibit strong orthogonal structure across different classes.}
By encouraging orthogonality with respect to teacher's predictive structure, student is guided toward a similar class-separating geometry, thereby facilitating teacher's inter-class structural knowledge transfer. BicKD realizes this objective by bilaterally amplifying orthogonality among distinct classes during distillation, promoting clearer separation of class probability directions while preserving consistency within the same class.

To measure the orthogonality in distillation, 
we employ the cosine distance $\mathbb{D}: \mathbb{R}^{n}\times \mathbb{R}^{n} \to \mathbb{R}^{+} $ define as
\begin{equation}
\mathbb{D}(\mathbf{u}, \mathbf{v})
:= 1 - \frac{\mathbf{u}^\top \mathbf{v}}
{\|\mathbf{u}\|_2 \, \|\mathbf{v}\|_2}.
\label{eq:cosDist}
\end{equation}
%where $\overline{u}$ and $\overline{v}$ are the means of vectors $\mathbf{u}$ and $\mathbf{v}$, respectively.
% In probability space, the cosine distance $\mathbb{D}_{\cos} \in [0, 1]$. 
%{\color{red}please check !!}
%\added[]{
Since the prediction vectors $\mathbf{u}, \mathbf{v}$ 
both lie in the probability simplex, their inner product satisfies $\mathbf{u}^\top\mathbf{v}\ge 0$ and 
their cosine similarity $\cos(\mathbf{u}, \mathbf{v})=\frac{\mathbf{u}^\top \mathbf{v}}
{\|\mathbf{u}\|_2 \, \|\mathbf{v}\|_2} \in [0, 1]$.
Therefore, the cosine distance $\mathbb{D}(\mathbf{u}, \mathbf{v})=1-\cos(\mathbf{u}, \mathbf{v}) \in [0, 1]$.
%generated by softmax(Eq. \ref{softmax}) are non-negative and sum to one, so both vectors lie in the probability simplex. Their inner product satisfies $\mathbf{u}^\top\mathbf{v}\ge 0$. Moreover, since neither vector is the zero vector, their cosine similarity $\cos(\mathbf{u}, \mathbf{v})=\frac{\mathbf{u}^\top \mathbf{v}}{\|\mathbf{u}\|_2 \, \|\mathbf{v}\|_2} \in [0, 1]$.
%
% Therefore, the cosine distance $\mathbb{D}=1-\cos(\mathbf{u}, \mathbf{v}) \in [0, 1]$.}
When $\mathbb{D}(\mathbf{u}, \mathbf{v}) = 1$, the inner product of $\mathbf{u}$ and $\mathbf{v}$ is $0$, which means two vectors are orthogonal. 
By maximizing this distance, orthogonality is amplified. 
Using cosine distance rather than cosine similarity more directly reflects our objective of encouraging orthogonality (i.e., dissimilarity), while also maintaining consistency with other distance-based terms in the objective function. 
%Moreover, using a distance-based formulation maintains consistency with other distance terms in the objective and provides a unified scale across loss components.
% By maximizing the distance, the orthogonality is amplified.

% By amplifying both the sample-wise and class-wise orthogonality, BicKD enables the strong classes to aid the learning of weak classes in distillation. 

\subsection{Sample-wise Contrastive Loss}
\label{sec:sample-contrast}
%\added[]{
The sample-wise contrast exploits geometric structure to guide student's predictive direction at the individual sample level. By emphasizing orthogonality, predictions from different classes are pushed further apart, while those belonging to the same class are pulled closer, resulting in a more discriminative probability geometry. Sample-wise alignment further ensures the correctness of individual predictions during distillation.%}

%%%%%%%%%%%%%%%%%%%%%%%%%%%%%%%%%%
\subsubsection{Sample-wise Orthogonality}
The sample-wise orthogonality is specifically for samples with different labels.
For each batch $\mathcal{B}$, denote the index set of sample pairs with distinct ground truth labels as 
 \begin{equation}\label{eq:deltaB}
     \Delta_{\mathcal{B}} := \{(i,j)\,|\,i,j \in \mathcal{B}, y_i \neq y_j\}.
 \end{equation}
Consider a sample pair with index $(i,j)\in \Delta_{\mathcal{B}}$. 
In the distillation process, BicKD aims to make 
the student's prediction $\mathbf{S}_{i,:}$ on sample $i$ orthogonal to the teacher's prediction $\mathbf{T}_{j,:}$ on sample $j$. %, where $i\neq j$. 
In other words, the distance $\mathbb{D}(\mathbf{S}_{i,:}, \mathbf{T}_{j,:})$ is expected to be as large as possible. 
Accordingly, the sample-wise orthogonality amplification (SOA) loss is defined as
{
\begin{equation}\label{eq:RDA}
    \mathcal{L}_{soa} := -\frac{1}{|\Delta_{\mathcal{B}}|} \sum\limits_{(i,j)\in \Delta_{\mathcal{B}}} \mathbb{D}(\mathbf{S}_{i,:}, \mathbf{T}_{j,:}).
\end{equation}}

\subsubsection{Sample-wise Alignment.}
\label{subsec:VKD}
To amplify orthogonality and enhance distillation, alignment is applied for the contrastive strategy.
Following vanilla KD \cite{KD}, 
we adopt 
Kullback-Leibler divergence based loss $\mathcal{L}_{KL}$
%$\mathcal{L}_{KL} := \frac{1}{B}\sum^B_{i=1}KL(\mathbf{S}_{i,:},\mathbf{T}_{i,:})$
for sample-wise alignment.

% $\mathcal{L}_{KL}$ for sample-wise alignment.
% employs unilateral sample-wise alignment to preserve the relations between teacher and student models via
% $\mathcal{L}_{KL} := \frac{1}{B}\sum^B_{i=1}KL(\mathbf{S}_{i,:},\mathbf{T}_{i,:})$
% where $KL$ refers to Kullback-Leibler divergence.
% \added[]{We adopt $\mathcal{L}_{KL}$ for sample-wise alignment.}

%%%%%%%%%%%%%%%%%%%%%%%%%%%%%%%%%%%%%%%%%

\subsection{Class-wise Contrastive Loss}
As discussed in Section \ref{sec:intro}, %{\color{red}which section??}, 
%discussed earlier, 
vanilla KD only employs \emph{unilateral} sample-wise alignment, yet it lacks an inter-class comparison mechanism, and therefore does not fully capture the structural relationships among different classes reflected in teacher’s output probabilities.

The sample-wise contrastive loss (Section \ref{sec:sample-contrast}) only performs contrast within samples in the current batch. To further capture class-level structure, we introduce class-wise relational constraints at a complementary granularity, which enables further cross-class comparison and promotes more faithful transfer of teacher's probability geometry.

\subsubsection{Class-wise Orthogonality}
  % The above sample-wise orthogonality only ensures that samples in the current batch are contrasted, which 
  % \added[]{limits the model's ability to fully exploit inter-class differences.}
% To ensure \added[]{the weak classes} receive adequate attention and learn from the strong classes' strengths, we introduce class-wise orthogonality amplification.  

In each batch, the column $\mathbf{T}_{:,k}$ represents 
teacher's predictions for  class $k$, and the column $\mathbf{S}_{:,j}$ indicates student's confidence in class $j$. 
For different classes $j\ne k$, the cosine distance between $\mathbf{S}_{:,j}$ and $\mathbf{T}_{:,k}$ is expected to be as large as possible. In other words, $\mathbf{S}_{:,j}$ and $\mathbf{T}_{:,k}$ is expected to be orthogonal.
Thus the class-wise orthogonality amplification (COA) loss is formulated as 
\begin{equation}\label{eq:CDA}
    \mathcal{L}_{coa} := -\frac{1}{C(C-1)} \sum\limits_{j \neq k} \mathbb{D}(\mathbf{S}_{:,j}, \mathbf{T}_{:,k}).
\end{equation}
By leveraging class-wise orthogonality amplification, the student is able to learn teacher's inter-class structural knowledge, thereby enhancing its generalization capability.

%%%%%%%%%%%%%%%%%%%%%%%%%%%%%%%%%%%%
\subsubsection{Class-wise Alignment}
We exploit a $L_1$-distance based class-wise alignment to narrow the discrepancies between teacher and student's predictions on identical classes.
%We exploit the $L_1$ distance to align their predictions on identical classes.
Accordingly the class-wise alignment (CA) loss is 
$$\mathcal{L}_{ca} := \frac{1}{C} \sum_{j=1}^C L_1(\mathbf{S}_{:,j}, \mathbf{T}_{:,j})$$
where $L_1(\cdot,\cdot)$ is the standard $L_1$ distance.

%%%%%%%%%%%%%%%%%%%%%%%
\begin{algorithm}[!htbp]
\caption{The proposed BicKD}\label{DAprogram}
\textbf{Input:} Dataset $\mathbf{X}$, student model $S$ and its parameters $\boldsymbol{\theta}$, teacher model $T$, label $Y$, hyperparameters $\alpha$, $\beta$, $\gamma$, and the initial learning rate $\eta_0$.\\
\textbf{Initialization:} 
Initialize student model $S$, index $t=0$.

\textbf{Repeat}\\
    \text{1:\ } Input a batch of $\mathbf{X}$ to teacher and student models, output their predictions $\mathbf{T}$, $\mathbf{S}$ \\
    \text{2:\ } Calculate cross-entropy loss $\mathcal{L}_{CE}$ \\
    \text{3:\ } Calculate sample-wise contrastive loss via ~\eqref{eq:SC} %$$\mathcal{L}_{SC} = \mathcal{L}_{soa} + \mathcal{L}_{KL}$$
    \\
    \text{4:\ } Calculate class-wise contrastive loss via ~\eqref{eq:CC} 
    %$$\mathcal{L}_{CC} = \mathcal{L}_{coa} + \mathcal{L}_{ca}$$
    \\
    \text{5:\ } Build BicKD total training loss via ~\eqref{eq:BicKD_s}
    %\begin{align*}
    %$$\mathcal{L}_{train} \gets \alpha \mathcal{L}_{CE} + \beta \mathcal{L}_{SC} + \gamma \mathcal{L}_{CC}$$
    \\
    %\end{align*}
    % \text{Step 6:} Update $lr_{epoch}$ \\
    \text{6:\ } Update $\eta_{t}$ and train student model $S$ by 
    $$\boldsymbol{\theta}_{t + 1} \gets \boldsymbol{\theta}_{t} - \eta_{t} \times \frac{\partial \mathcal{L}_{train}}{\partial \boldsymbol{\theta}_{t}}.$$
    
\textbf{Until} Converge  

\textbf{Return} the student model $S$
\end{algorithm}
%%%%%%%%%%%%%%%%%%%%%%%%%%%%%%%%

%%%%%%%%%%%%%%%%%%%%%%%%%%%%%%%%%%%%%%%%%%

%%%%%%%%%%%%%%%%%%%%%%%%%%%%%%%%%%%%%%%%
%%%%%%%%%%%%%%%%%%%%%%%%%%%%%%%%%%%%%%%%%%%%%%%%%%%%%%%%%%%%%%%%%%%%%%%%
\begin{table*}[!htbp]%\small
    \centering
    \tabcolsep=0.45cm
    \caption{Accuracy (\%) of BicKD and baselines on CIFAR-100.}
    \begin{tabularx}{\linewidth}{c c c c c c c c c c c}
        \toprule
        Teacher  &     \multicolumn{2}{c}{ResNet-56}&\multicolumn{2}{c}{WRN-40-2}&\multicolumn{2}{c}{ResNet-110} &\multicolumn{2}{c}{VGG13}   &  \multicolumn{2}{c}{ShuffleNetV2} \\
        Student   &     \multicolumn{2}{c}{ResNet-20}&\multicolumn{2}{c}{WRN-16-1}&\multicolumn{2}{c}{WRN-16-2} &\multicolumn{2}{c}{ResNet-20}   &  \multicolumn{2}{c}{ResNet-14} \\
        \midrule
        Teacher  &\multicolumn{2}{c}{72.42}&\multicolumn{2}{c}{75.59}&\multicolumn{2}{c}{74.37}&\multicolumn{2}{c}{74.64}&\multicolumn{2}{c}{72.80} \\
        Student  &\multicolumn{2}{c}{69.06}&\multicolumn{2}{c}{66.74}&\multicolumn{2}{c}{73.26}&\multicolumn{2}{c}{69.06}&\multicolumn{2}{c}{66.41} \\
        \midrule
        &     top-1 &top-5 &top-1 &top-5 &top-1 & top-5 & top-1 & top-5  & top-1 &top-5   \\
        \cmidrule(lr){2-3} \cmidrule(lr){4-5} \cmidrule(lr){6-7} \cmidrule(lr){8-9}  \cmidrule(lr){10-11}  
        % \cmidrule(lr){12-13} \cmidrule(lr){14-15}
        KD \cite{KD}&     70.43&92.34
        &66.86&90.78
        &73.88& 93.57
        & 69.72 & 91.70  & 66.49& 90.35
        \\
           VID \cite{VID} &     69.40&91.93
        &66.92&90.54
        &72.04& 92.62
        & 69.73 & 91.80  & 65.83& 90.06
        \\
        FitNets \cite{fitnets}&     67.93&91.04
        &66.82&90.80
        &69.63& 91.32
        & 69.31 & 91.70  & 65.04& 89.65
        \\
           RKD \cite{relationalKD} &     68.64&91.52
        &66.01&90.06
        &71.21& 92.30
        & 69.01 & 91.46  & 65.10& 89.86
        \\
            SP \cite{SP} &     69.16&91.92
        &66.76&90.82
        &71.39& 92.15
        & 68.67 & 91.82  & 65.63& 90.17
        \\
           CRD \cite{CRD} &     69.91&92.05
        &67.64&\textbf{91.19}
        &72.59& 92.82
        & 69.78 & 91.95  & 66.22& 90.38
        \\
          DIST \cite{dist} &     69.40&91.42
        &67.55&90.61
        &70.92& 91.95
        & \underline{69.97} & \underline{91.98}  & 66.82& 90.06
        \\
          MLD \cite{multi-level-logits}  &     70.33&92.19
        &\underline{67.71}&90.90
        &\underline{74.23}& \textbf{93.69}
        & 69.83 & 91.60  & 67.13& 90.58
        \\
          RLD\cite{rld2024} &     69.50&92.29
        &66.62&91.06
        &73.16& 93.43
        & 69.48 & \textbf{91.99}  & 65.28& 90.70
        \\
          WKD-L\cite{wkd2024} &     70.29&92.34
        &67.47&91.12
        &73.21& 93.14
        & 69.66 & 91.54  & 67.11& \textbf{91.08}
        \\
          WTTM\cite{wttm2024} &     \underline{70.93}&\underline{92.48}
        &67.46&90.99
        &73.94& 93.59
        & 69.14 & 91.50  & \underline{67.17}&\underline{90.88}
        \\
         \textbf{BicKD}&     \textbf{71.85}&\textbf{92.61}
        &\textbf{68.63}&\underline{91.16}
        &\textbf{75.15}& \underline{93.68}& \textbf{70.47}& 91.58& \textbf{68.76}&90.80\\
        \bottomrule
    \end{tabularx}
    \label{tab:cifaracc}
\end{table*}
%%%%%%%%%%%%%%%%%%%%%%%%%%%%%%%%%%%%%

%%%%%%%%%%%%%%%%%%%%%%%%%%%%%%%%%%%%%%%%%%
\subsection{Bilateral Contrastive Knowledge Distillation}
  Combining the alignment and orthogonality amplification, the sample-wise contrastive loss (SC) is 
\begin{align}\label{eq:SC}
   \mathcal{L}_{SC} := \mathcal{L}_{soa} + \mathcal{L}_{KL}, 
\end{align}
and the class-wise contrastive loss (CC) is
\begin{align}\label{eq:CC}
    \mathcal{L}_{CC} := \mathcal{L}_{coa} + \mathcal{L}_{ca}.
\end{align}
In the distillation, 
BicKD exploits an integration of the ground truth labels $\mathcal{L}_{CE}$, sample-wise contrast $\mathcal{L}_{SC}$ and class-wise contrast $\mathcal{L}_{CC}$ to achieve bilateral contrast from teacher to student models. 
The overall BicKD training loss is
\setlength\abovedisplayskip{0.2cm}
\setlength\belowdisplayskip{0.2cm}
\begin{align}\label{eq:BicKD_s}
    \mathcal{L}_{train} &:= \alpha \mathcal{L}_{CE} + \beta \mathcal{L}_{SC} + \gamma \mathcal{L}_{CC}
\end{align}
where $\alpha$, $\beta$ and $\gamma$ are positive parameters balancing the three aspects (see also Algorithm \ref{DAprogram}).

In summary, the proposed BicKD 
% addresses the challenge of \added[]{the class-wise accuracy discrepancy}
provides the following advantages:

\begin{itemize}
    \item Simplicity. BicKD only requires logits from teacher and student models, without the need for feature representations or data augmentations, making it simple, cost-effective, and easy to implement.
    \item Comprehensiveness. 
    BicKD contrasts not only sample-wise but also class-wise.
    By amplifying the orthogonality among different classes, BicKD enables student model to learn teacher’s geometric structure in probability space in terms of cross-class comparison, resulting in more effective  and discriminative knowledge transfer.
    % enhances the student model's learning capability for weak classes by leveraging the strengths of strong classes. 
    % This is highly advantageous for \added[]{weak classes with lower learnability.}
\end{itemize}

\section{Experiments}
\label{sec:experiments}
\subsection{Experiment Settings}
\label{sec:exsetting}
  \textbf{Datasets.} 
We conduct image classification tasks using two typical datasets: 
1) CIFAR-100 \cite{CIFAR100} consist of 50K RGB training images and 10K testing images across 100 classes, respectively, with a resolution of 32$\times$32.
2) Tiny-ImageNet \cite{tinyimagenet} 
% used in the experiments is rearranged by the method \cite{mlclf}, to
encompass 50K RGB training images and 5K testing images across 200 classes, each sized 64$\times$64 pixels.

\textbf{Models.}
We evaluate the performances of KD methods on different architectures, including VGG \cite{vgg}, ResNet \cite{ResNet}, Wide ResNet \cite{wrn}, 
ResNeXT \cite{resnext}, 
ShuffleNetV2 \cite{shufflenetV2}
% ShuffleNetV1,V2 \cite{shufflenetV1,shufflenetV2}.
, and MobileNet \cite{mobilenets}.

\textbf{Implementation.}
To ensure equitable comparisons, we follow the pre-trained models and training settings as described in \cite{CRD}. 
We implement the BicKD loss \eqref{eq:BicKD_s}, determining the optimal hyperparameters  by conducting a grid search: $\alpha = 1$, $\beta=2$, $\gamma=2$, within the ranges of $\alpha$ from 0.5 to 2, $\beta$ from 1 to 2, and 
$\gamma$ from 1 to 2. 
For the vanilla KD loss \eqref{eq:KD}, we adopt $\lambda = 0.1$ as recommended by \cite{KD}. 

We implement experiments with a batch size 256 and the total epoch number 240.
We apply stochastic gradient descent with Nesterov momentum of $0.9$ and weight decay of $0.0005$ as the optimizer.
The initial learning rate is $0.01$ for ShuffleNet and MobileNet, and $0.05$ for other models.
The learning rate is decayed by a factor of $0.1$ at epoch 150, 180, 210.
The temperature parameter $\tau$ used in KL divergence is set to be 4.

The hardware used in our experiments includes two PCs with GeForce RTX 4090 GPU and Intel(R) Core(TM) i9-10900X CPU @ $3.70$GHz CPU.

\textbf{Baselines.}
We compare BicKD with existing baselines illustrated in Table ~\ref{tab:baselines}.
% All methods are %trained on standard datasets, and 
% evaluated on test datasets 
% by Top-1 and Top-5 accuracy.
To ensure consistency and reduce variability, %we fix the random seed to $1$ and 
the accuracies reported in this paper are the average results from three repeated experiments.
% \\[-0.6cm]

\textbf{Pre-trained Models.}
The teacher models are pre-trained on a complete CIFAR-100 or Tiny-ImageNet dataset using only the ground truth labels.

\subsection{Image Classification on CIFAR-100}
We perform experiments on
standard CIFAR-100 dataset to evaluate the effectiveness of BicKD and other baseline methods using five distinct distillation pairs, including both homogeneous and heterogeneous settings.
The top-1 and top-5 \replaced[]{test}{validation} accuracy (\%) comparison results of BicKD and other distillation approaches are reported in Table ~\ref{tab:cifaracc}.
It is readily seen that our BicKD method surpasses other baseline methods, showing its effectiveness to capture class-wise differences in generalization and strengthen student model's overall discriminative capacity. 
In particular, when teacher is ResNet-110, the student WRN-16-2 distilled by BicKD (75.15\%) surpasses teacher model (74.37\%) and state-of-the-art baselines, showing the superiority of BicKD.
%, affirming its superiority.
%\added[]{%and using only ground-truth labels, 
Under the same training protocol, BicKD achieves a stable average improvement 1.28\% over the baseline methods, with gains reaching up to 1.77\% in certain settings. These improvements indicate that explicitly modeling probability-space geometry and cross-class relationships plays an effective role in enhancing distillation performance.
%}

We further conduct statistical significance test comparing the accuracy of vanilla KD and BicKD.
% When $\delta=1$, 
As a result, BicKD significantly enhances student model's accuracy over vanilla KD, yielding an average p-value of 0.03736, which is below the significance threshold of $0.05$. 
%In other words, BicKD significantly outperforms the existing baselines on CIFAR-100.
%\added[]{standard} CIFAR-100 dataset.

%%%%%%%%%%%%%%%%%%%%%%%%%%%%%%%%%%%
\subsection{Image Classification on Tiny-ImageNet}
To further examine the scalability of the proposed BicKD on higher-resolution datasets, we conduct experiments on the Tiny-ImageNet dataset.
We report the Top-1 accuracy of student models under three different distillation pairs: ResNet-50$\to$WRN-40-1, ResNet-32x4$\to$WRN-16-2 and WRN-40-2$\to$ResNet-8.

As shown in Table~\ref{tab:tinyacc}, 
our BicKD method consistently achieves the highest Top-1 accuracy across all three distillation pairs. \replaced[]{The student ResNet-8 distilled by BicKD (65.28\%) also performs better than the teacher model WRN-40-2(62.84\%) and outperforms the second best method CRD
(64.80\%) by 0.48\%, showing the strong transfer capability of BicKD.}{The student ResNet-8 distilled by BicKD (65.28\%) also performs better than teacher model WRN-40-2 (62.84\%), showing the strong transfer capability of BicKD.} Besides, BicKD indicates stable improvements over competitive baselines, with an average gain of 0.49\% on Tiny-ImageNet. This shows that learning geometric structure and modeling cross-class relationships in BicKD are effective, leading to stable performance improvements.

In addition, a statistical significance test shows that, compared to vanilla KD, BicKD significantly improves the student model’s accuracy with an average p-value of 0.03733, below the significance threshold of 0.05.

\begin{table}[t]
    \centering
    \tabcolsep=0.24cm
    \caption{ Accuracy (\%) of BicKD and baselines on Tiny-ImageNet}
    \begin{tabular}{c c c c}
    \toprule
    Teacher& ResNet-50 & ResNet-32x4&WRN-40-2\\
    Student&WRN-40-1 & WRN-16-2& ResNet-8 \\
    \midrule
    Teacher &69.09  &63.32 &62.84 \\
    Student &54.64  &58.00 &60.45 \\
    % Teacher& \multicolumn{2}{c}{ResNet-50} & \multicolumn{2}{c}{ResNet-32x4}  \\
    % Student& \multicolumn{2}{c}{WRN-40-1} & \multicolumn{2}{c}{WRN-16-2}  \\
    \midrule
   %      & top-1 & top-1 & top-1 \\
   % % & top-5 \\
   %  % \cmidrule(lr){2-3} \cmidrule(lr){4-5} 
   %   \midrule
    KD \cite{KD}& 56.20&  58.07&63.07 \\
   VID \cite{VID} & 56.48& 57.58&62.25 \\
FitNets \cite{fitnets}& 55.90&  57.33&60.40 \\
   RKD \cite{relationalKD} & 56.28& 57.67&61.47\\
    SP \cite{SP} & 55.38& 58.58&64.04 \\
   CRD \cite{CRD} & \underline{58.37}& \underline{59.83}&\underline{64.80} \\
  DIST \cite{dist} & 57.57& 59.40&63.00 \\
  MLD \cite{multi-level-logits}  & 56.36& 59.16&63.34 \\
  RLD\cite{rld2024} & 57.50& 58.72&63.63\\
  WKD-L\cite{wkd2024} & 56.54& 58.80&64.26 \\
  WTTM\cite{wttm2024} & 57.25& 59.80&64.76 \\
    \textbf{BicKD} & \textbf{58.77}& \textbf{60.41}&\textbf{65.28} \\
    \bottomrule
    \end{tabular}
    \label{tab:tinyacc}
\end{table}

%%%%%%%%%%%%%%%%%%%%
\subsection{Discussion}
According to the experimental results  in Tables \ref{tab:cifaracc} and \ref{tab:tinyacc}, BicKD shows a consistent performance advantage under diverse teacher-student configurations. The gains are observed across architectures and become more pronounced under large structural gaps, such as ShuffleNetV2$\to$ResNet-14 and WRN-40-2$\to$ResNet-8. This implies that BicKD learns the geometric structure of teacher’s probability space rather than merely mimicking prediction outputs. These results also show the effectiveness of both sample-wise and class-wise contrast mechanisms in facilitating structured knowledge transfer.
%{\color{red}add more???}

%%%%%%%%%%%%%%%%%%%%%%%%%%%%%%%%%%%%%%%%%%%%%%%%%%%%%%%%%%%%%%%%%%%%%%%%%%%%%%%%%%%%%%%%
\section{Ablation studies}
\label{sec:ablationstudy}
\subsection{Unilateral Contrast}
\label{sec:uncon}
  BicKD consists of two components: sample-wise contrast and class-wise contrast.
  To verify the effectiveness of bilateral contrast in BicKD, we assess the performance of sample-wise or class-wise contrastive distillation alone, denoted as SC and CC respectively\added[]{, where the distillation loss of BicKD is replaced by solely \eqref{eq:SC} or \eqref{eq:CC}.}
  Table \ref{tab:BicAbla} illustrates the experimental results of top-1 \added[]{and top-5} accuracy on CIFAR-100.
%As Table ~\ref{tab:BicAbla} illustrated, 
It is observed that each individual SC or CC contrast strategy outperforms vanilla KD. BicKD further improves performance by jointly integrating the two components.

\begin{table}[h]
    \centering
    \caption{Ablation study of sample-wise and class-wise contrast on CIFAR-100.}
    \begin{tabular}{c c c c c}
    \toprule
    Teacher$\to$Student& \multicolumn{2}{c}{ResNet-50$\to$ResNet-8}& \multicolumn{2}{c}{ShuffleNetV2$\to$ResNet-8}\\
    \midrule
    &top-1& top-5& top-1&top-5\\
    \cmidrule(lr){2-3} \cmidrule(lr){4-5} 
    KD\cite{KD} & 58.26 & 86.24  & 57.64 &85.42  
\\
    SC & 61.07 & \underline{87.38}   & 61.27&\underline{87.21}\\
    CC & \underline{61.37} & 86.35   & \underline{61.63}&86.56
\\
 \textbf{BicKD} & \textbf{62.18} & \textbf{87.55}  & \textbf{62.35}&\textbf{87.67}
\\
 \bottomrule
    \end{tabular}
    \label{tab:BicAbla}
\end{table}

\begin{table*}[h]
    \centering
    \caption{Per-batch training time of different knowledge distillation methods.}
    \begin{tabular}{ccccccccccccc}
    \toprule
       Method & KD & VID & FitNets & RKD & SP & CRD & DIST & MLD & RLD & 
        WKD-L & WTTM & \textbf{BicKD} \\
        \midrule
        feature/logits-based& logits & feature & feature & feature& feature & feature & logits & logits & logits & 
        logits & logits & logits \\
        % \midrule
        reference 
        & \cite{KD} & \cite{VID} & \cite{fitnets} & \cite{relationalKD} & \cite{SP} & \cite{CRD} 
        & \cite{dist} & \cite{multi-level-logits}
        & \cite{rld2024} 
        & \cite{wkd2024}&\cite{wttm2024} 
        & this work \\
        \midrule
        \makecell{CIFAR-100 \\ ResNet-56$\to$ResNet-20} &  0.0461&  0.0526&  0.0459&  0.0516& 0.0453&  0.1752&  0.0460&  0.0485& 0.0454& 0.0493&\underline{0.0452}&\textbf{0.0450}\\
        \midrule
        \makecell{Tiny-ImageNet \\ WRN-40-2$\to$ResNet-8} &  0.0783&  0.0781&  0.0783&  0.0785& 0.0791&  0.3508&  \textbf{0.0773}&  \underline{0.0778}& 0.0781& 0.0779&0.0784&\underline{0.0778}\\
        \bottomrule
    \end{tabular}
    \label{tab:time}
\end{table*}

\subsection{Unilateral Orthogonality Amplification}
%  The BicKD amplifies the orthogonality from two aspects: sample-wise and class-wise orthogonality. 
To verify the effectiveness of orthogonality amplification, 
we conduct experiments to assess the performance of individually sample-wise or class-wise orthogonality amplified, denoted as OA(S) and OA(C) respectively. 
\added[]{Specifically, OA(S) and OA(C) are derived from SC and CC respectively by removing the same-class alignment term, retaining only the orthogonality amplification mechanism, where the distillation loss is replaced by solely \eqref{eq:RDA} or \eqref{eq:CDA}.}
Table \ref{tab:ablation} illustrates the experimental top-1 and top-5 accuracy
on CIFAR-100. 
The results show that amplifying the orthogonality on any single aspect can surpass vanilla KD. Further, BicKD boosts the performance by combining the two aspects together.  

\begin{table}[t]
    \centering
    \caption{Ablation study of sample-wise or class-wise orthogonality amplification on CIFAR-100.}
    \begin{tabular}{c c c c c}
    \toprule 
  Method& \multicolumn{2}{c}{ResNet-50 $\to$ ResNet-8} & \multicolumn{2}{c}{WRN-40-2 $\to$ ResNet-14}\\
 \midrule
 &  top-1&top-5& top-1&top-5\\
    % Teacher & ResNet-50& WRN-40-2\\
    % Student & ResNet-8& MobileNetV2\\
    \cmidrule(lr){2-3} \cmidrule(lr){4-5}
 KD \cite{KD} & 58.26           &86.24& 66.44&90.61
\\
OA(S)& \underline{59.78}           &\underline{86.80} & \underline{67.08}&90.72
\\
OA(C)& 59.47           &86.40& 66.97&\underline{90.73}
\\
\textbf{BicKD} & \textbf{62.18}  &\textbf{87.55}& \textbf{68.79}&\textbf{91.22}
\\
    \bottomrule
    \end{tabular}
    \label{tab:ablation}
\end{table}

\subsection{BicKD on Few-Shot Dataset}

\begin{figure}[t]
    \centering    
    \includegraphics[width=\linewidth]{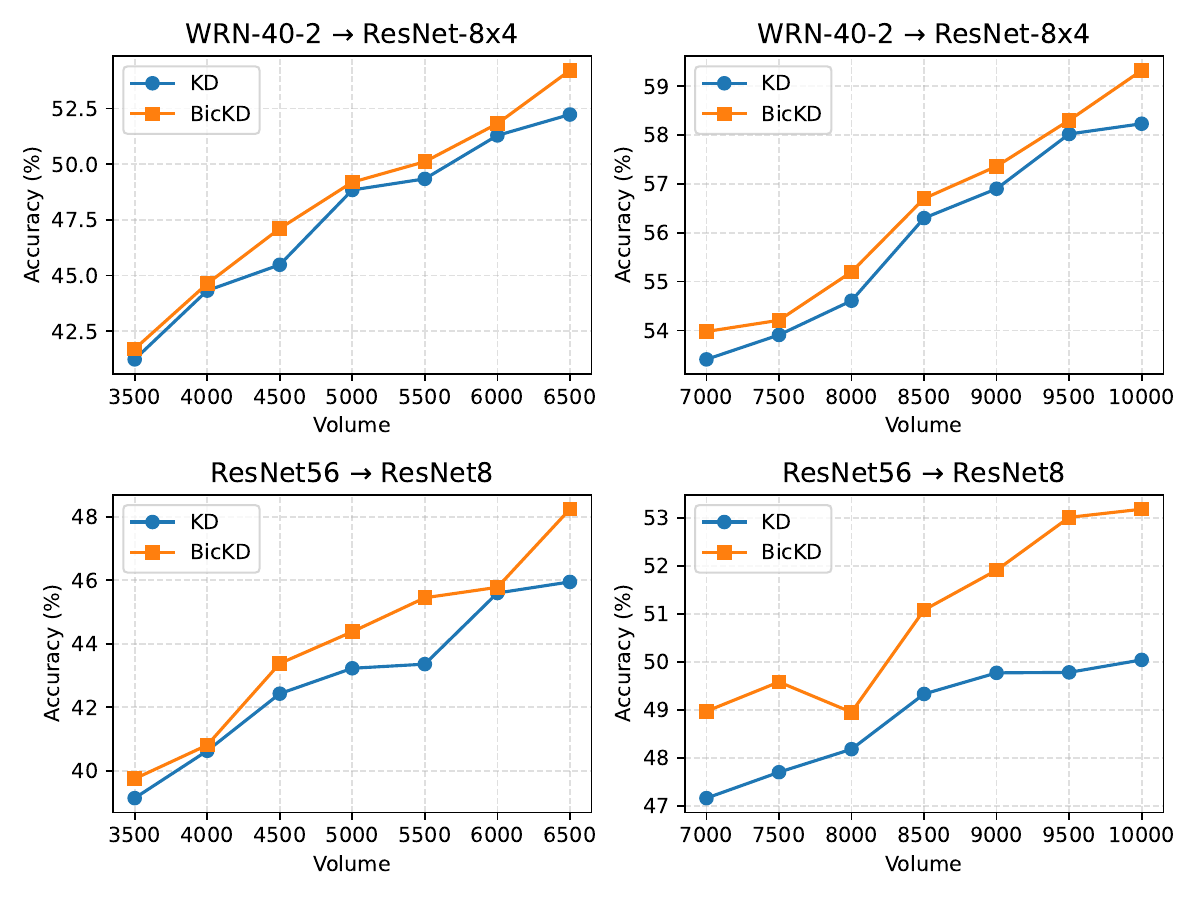}
    \caption{Accuracy (\%) of BicKD and Vanilla KD on CIFAR-100-FS with varying data volumes.
 }
    \label{fig:fs}
\end{figure}

We conduct experiments on an extreme few-shot dataset -- CIFAR-100-FS, which contains very few images for each class. Following \cite{cifarfs}, we randomly select between 35 and 100 samples for each class from the CIFAR-100 dataset, resulting in a transfer dataset ranging from 3500 to 10000, with a step size of 500. We evaluate the performance of BicKD and vanilla KD using two  teacher and student model pairs: (WRN-40-2$\xrightarrow{}$ResNet-8x4) and (ResNet-56$\xrightarrow{}$ResNet-8).

As shown in Fig.~\ref{fig:fs}, BicKD consistently outperforms vanilla KD across all dataset volumes.
For instance, when ResNet-56$\xrightarrow{}$ResNet-8 and volume 7000, BicKD achieves an accuracy of $48.97\%$ while vanilla KD is only $47.16\%$.
Notably, as the dataset volume decreases, the performance gap between BicKD and vanilla KD remains significant, which indicates BicKD's superiority in few-shot learning scenarios.

\subsection{BicKD on Long-tailed Dataset}
Table~\ref{tab:BicAbla-LT} presents the empirical results of BicKD and vanilla KD on a long-tailed dataset CIFAR-100-LT \cite{cifarlt} sampled from CIFAR-100 with an imbalance factor $\rho=10, 100$, %We set the $\rho$ as 10 and 100. 
using two different teacher and student model pairs: (ResNet-50$\to$ResNet-8)
and (WRN-40-2 $\to$ MobileNetV2).
\begin{table}[h]
    \centering
    \tabcolsep=0.14cm
    \caption{Ablation study of sample-wise and class-wise contrast on CIFAR-100-LT.}
    \begin{tabular}{c c c c c c c}
    \toprule
    Method& \multicolumn{3}{c}{ResNet-50 $\to$ ResNet-8} & \multicolumn{3}{c}{WRN-40-2 $\to$ MobileNetV2
}\\
    \midrule
    &{\small *} & {\small $\rho=10$} & {\small $\rho=100$}  & {\small *} & {\small $\rho=10$} &{\small $\rho=100$}  
\\
    KD\cite{KD} & 58.26 & 46.23& 32.39& 61.35& 43.74&28.17\\
 BicKD & \textbf{62.18} & \textbf{50.50}& \textbf{36.32}& \textbf{63.47}& \textbf{44.97}&\textbf{30.92}\\
    \bottomrule
    \end{tabular}
    \label{tab:BicAbla-LT}
\end{table}

Moreover, we evaluate the improvement of BicKD on minority classes with very few samples
in CIFAR-100-LT.

\begin{table}[h]
    \centering
    \tabcolsep=0.1cm
    \caption{Accuracy (\%) on 10 minority  classes with least number of samples in ResNet-50 $\to$ ShuffleNetV2 on CIFAR-100-LT with $\rho=100$.}
    \begin{tabular}{c c c c c c c c c c c}
    \toprule
    \multicolumn{11}{c}{Minority  classes in CIFAR-100-LT with $\rho=100$}\\
    \midrule
Num. of samp. & 5 & 5 & 5 & 5 & 6 & 6 & 6 & 6 & 7 & 7 \\
\midrule
    KD \cite{KD} & 4.0& 1.0& \textbf{5.0}& 8.0& 10.0& 29.0& 5.0& 6.0& 15.0 & 13.0\\
        \textbf{BicKD} & \textbf{10.0}& \textbf{2.0} & 2.0& \textbf{10.0}& \textbf{16.0}& \textbf{44.0}& \textbf{6.0}& \textbf{8.0}& \textbf{23.0}& \textbf{14.0}\\
    \bottomrule
    \end{tabular}
    \label{tab:classwise}
\end{table}

% \begin{table*}[!htbp]
%     \centering
%     % \tabcolsep=0.25cm
%     \caption{Per-batch runtime of different knowledge distillation methods}
%     \begin{tabular}{c c c c c c c c c c c}
%     \toprule
%     Method & FitNets & KD & VID & SP & CRD & DIST & MLD & RLD & 
%     WTTM & WKD-L & \textbf{BicKD} \\
%     \midrule
%     feature/logits-based 
%     & feature & logits & feature & feature & feature & logits & logits & logits & 
%     logits & logits & logits \\
%     \midrule
%     reference 
%     & \cite{fitnets} & \cite{KD} & \cite{VID} & \cite{SP} & \cite{CRD} 
%     & \cite{dist} & \cite{multi-level-logits}
%     & \cite{rld2024} & 
%     \cite{wttm2024} & \cite{wkd2024}
%     & This work \\
%     \bottomrule
%     \end{tabular}
%     \label{tab:time}
% \end{table*}

Table~\ref{tab:classwise} illustrates the accuracy of 10 least minority classes in 
% two scenarios:
% 1) (ResNet-32x4 $\to$ ResNet-8x4) on CIFAR-100-Dir with $\delta=0.1$, and 2) 
ResNet-50 $\to$ ShuffleNetV2 on CIFAR-100-LT with $\rho=100$.
It is readily seen that BicKD improves the performance on almost every minority classes. 
In particular, BicKD doubles the accuracy on two classes with only 5 samples, and also brings consistent improvements for classes containing six or seven samples, indicating the significant advantage of BicKD.
BicKD also enhances consistency among all classes, 
as indicated by the standard deviations: $27.03$ for BicKD compared to $27.39$ for vanilla KD on CIFAR-100-LT.

\subsection{Runtime Analysis}
\added[]{To demonstrate the efficiency of BicKD, we report the per-batch training time (in seconds) under two different teacher-student pairs: (ResNet-56 $\to$ ResNet-20) on CIFAR-100 and (WRN-40-2 $\to$ ResNet-8) on Tiny-ImageNet.
Notably, BicKD achieves the highest test accuracy among all methods in both settings. 
As shown in Table \ref{tab:time}, our method BicKD achieves a per-batch training time of 0.0450s on CIFAR-100 and 0.0778s on Tiny-ImageNet, showing its lightweight nature.
These results validate that BicKD achieves superior accuracy while maintaining computational efficiency.
}
% , comparable to vanilla KD (0.0783s) and faster than most feature-based baselines. 
% Besides, as a logit-based method, BicKD requires no additional memory overhead. 
% 

\section{Conclusion}
\label{sec:conclusion}
This paper \replaced[]{proposes}{proposed} a new method BicKD by introducing a bilateral contrast that amplifies orthogonality both sample-wise and class-wise. 
It introduces cross-class comparison mechanisms that are absent in vanilla KD, and imposes orthogonality constraints to transfer the geometric structure in teacher's probability space. %Specifically, 
% This paper investigates 
% \added[]{class-wise accuracy differences in Knowledge distillation.}
% We first empirically evaluated the performance of existing KD methods.  
% \added[]{The generalization ability differs across classes, resulting in large accuracy differences that negatively affect the overall model accuracy. }
% In an effort to mitigate it, 
By emphasizing orthogonality, the student is encouraged to learn the geometric structure from \added[]{the} teacher, which is shown in our experiments to be beneficial for knowledge distillation. The proposed bilateral contrast enables both sample-wise comparison and \deleted[]{introduces }cross-class knowledge, allowing the model to capture richer inter-class relationships.
Extensive experiments show that our BicKD outperforms the state-of-the-art distillation methods on different datasets.
Ablation analyses further confirm the contribution of each component and highlight the strong generalization capability of BicKD.

As future work, it is of interest to extend the orthogonality analysis to high-dimensional feature space, and to further explore the relationship between orthogonality and the simplex equiangular tight frame (ETF) structure in feature representations within the context of Neural Collapse~\cite{NCpapyan2020}.
% Simplex Equiangular Tight Frame
% \section*{Acknowledgment}

% The preferred spelling of the word ``acknowledgment'' in America is without 
% an ``e'' after the ``g''. Avoid the stilted expression ``one of us (R. B. 
% G.) thanks $\ldots$''. Instead, try ``R. B. G. thanks$\ldots$''. Put sponsor 
% acknowledgments in the unnumbered footnote on the first page.

\bibliographystyle{IEEEtran}
\balance
\bibliography{ref}
% \bibitem{b1} G. Eason, B. Noble, and I. N. Sneddon, ``On certain integrals of Lipschitz-Hankel type involving products of Bessel functions,'' Phil. Trans. Roy. Soc. London, vol. A247, pp. 529--551, April 1955.
% \bibitem{b2} J. Clerk Maxwell, A Treatise on Electricity and Magnetism, 3rd ed., vol. 2. Oxford: Clarendon, 1892, pp.68--73.
% \bibitem{b3} I. S. Jacobs and C. P. Bean, ``Fine particles, thin films and exchange anisotropy,'' in Magnetism, vol. III, G. T. Rado and H. Suhl, Eds. New York: Academic, 1963, pp. 271--350.
% \bibitem{b4} K. Elissa, ``Title of paper if known,'' unpublished.
% \bibitem{b5} R. Nicole, ``Title of paper with only first word capitalized,'' J. Name Stand. Abbrev., in press.
% \bibitem{b6} Y. Yorozu, M. Hirano, K. Oka, and Y. Tagawa, ``Electron spectroscopy studies on magneto-optical media and plastic substrate interface,'' IEEE Transl. J. Magn. Japan, vol. 2, pp. 740--741, August 1987 [Digests 9th Annual Conf. Magnetics Japan, p. 301, 1982].
% \bibitem{b7} M. Young, The Technical Writer's Handbook. Mill Valley, CA: University Science, 1989.
% \end{thebibliography}

\end{document}